# Using a Classifier Ensemble for Proactive Quality Monitoring and Control: the impact of the choice of classifiers types, selection criterion, and fusion process

Philippe THOMAS[a*], Hind BRIL EL HAOUZI[a], Marie-Christine SUHNER[a], André THOMAS[a], Emmanuel ZIMMERMANN[a,b], Mélanie NOYEL[a,b]

[a]*Université de Lorraine, CRAN, UMR 7039 CNRS, Campus Sciences, BP 70239, 54506 Vandœuvre-lès-Nancy cedex, France*

[b]*ACTA Mobilier Parc d'activité Macherin Auxerre Nord, 89470 Monéteau, France*

[*]*Corresponding author*

Phone: +33 (0)3 72 74 53 01

Fax: +33 (0)3 72 74 53 08

Email: philippe.thomas@univ-lorraine.fr; hind.el-haouzi@univ-lorraine.fr; Marie-Christine.Suhner@univ-lorraine.fr; andre.thomas@univ-lorraine.fr; ezimmermann@acta-mobilier.fr; mnoyel@acta-mobilier.fr

**Abstract.** In recent times, the manufacturing processes are faced with many external or internal (the increase of customized product re-scheduling, process reliability,.. ) changes. Therefore, monitoring and quality management activities for these manufacturing processes are difficult. Thus, the managers need more proactive approaches to deal with this variability. In this study, a proactive quality monitoring and control approach based on classifiers to predict defect occurrences and provide optimal values for factors critical to the quality processes is proposed. In a previous work (Noyel et al. 2013), the classification approach had been used in order to improve the quality of a lacquering process at a company plant; the results obtained are promising, but the accuracy of the classification model used needs to be improved. One way to achieve this is to construct a committee of classifiers (referred to as an ensemble) to obtain a better predictive model than its constituent models. However, the selection of the best classification methods and the construction of the final ensemble still poses a challenging issue. In this study, we focus and analyze the impact of the choice of classifier types on the accuracy of the classifier ensemble; in addition, we explore the effects of the selection criterion and fusion process on the ensemble accuracy as well. Several fusion scenarios were tested and compared based on a real-world case. Our results show that using an ensemble classification leads to an increase in the accuracy of the classifier models. Consequently, the monitoring and control of the considered real-world case can be improved.

# 1 Introduction

The growing need for complex and customized products and services has led to increased complexity of the associated manufacturing processes as well; this complexity may be attributed to several different sources depending on the features of the product/service and the organizational structure of the companies. Consequently, the manufacturing and control tasks become difficult, including their monitoring and quality management. Despite the many methods and operational tools that have been developed in the last few decades, the executive management personnel of these companies are always seeking new approaches and tools to analyze their specific problems and devise potential improvement strategies at several levels. These approaches and tools can thus be regarded as decision-aiding tools to identify not only the root causes of defects but also factors critical to the quality of their products/services; in particular, the aim of the executive managers is to eliminate those causes or limit their impact by setting the factors critical to quality at adequate or optimal levels. Among these approaches, a common method is the Design of Experiments (DoE); however, the primary disadvantage of such an approach is that the improvement process is considered "off-line." Indeed, even if a robust process is established by successfully optimizing the controllable and uncontrollable factors, this approach remains static, unable to take into account large and lumpy variations of all related factors throughout the process life cycle. Therefore, to handle these variations, online monitoring approaches are required. Moreover, in modern industries, many companies have adopted digital transformation, facilitated by recent technological advances in the field of communication and computer science, consequently, leading to the generation of large amounts of data from the manufacturing processes and from different assets, such as machines, products, and plants; this poses a challenge to utilize these data dynamically to monitor and manage the quality of these processes. Thus, machine learning is considered a suitable alternative to address this situation.

Furthermore, this paper reports on a study conducted in collaboration with a furniture company faced with some critical quality problems owing to the complexity of product flows because of considerably customized products with different routing sheets, as well as the complexity of some manufacturing processes, many of which are based on uncontrollable factors, including temperature, pressure, and their interactions. Previous research (Noyel et al. 2016) has demonstrated the significant advantages of an online quality monitoring approach based on defects classification using the neural network model; in addition, the research has highlighted some interesting perspectives, for example, improving the accuracy of the classifiers used. The primary aim of our study is to improve the accuracy of these classifiers. In this study, we primarily consider continuous data, and in this context, the most suitable classifier types are logic-based algorithms, neural networks, instance approaches, and support vector machines (SVMs). Nevertheless, these tools lead to classifiers with varying performances. Considering this, two approaches could be used: either selecting the classifier that yields the best results on a validation dataset or constructing a committee of classifiers to take advantage of the diversity of combined classifiers; the second approach is based on the hypothesis that a committee of classifiers, in general, outperforms its members (Kittler et al. 1998). This committee of classifiers is known by several names such as committees of learners, mixture of experts, classifier ensembles, and multiple classifier systems (Wozniak et al. 2014).

Classifier ensembles are often built using only one type of classifier; for example, neural network (NN) ensembles (Zhou et al. 2002, Windeatt 2005), SVM ensembles (Hachichi et al. 2011, Li et al. 2012), k-nearest neighbor (kNN) ensembles (Kim and Oh 2008, Ko et al. 2008), or tree ensembles (Dietterich 2000, Tsoumakas et al. 2009, Soto et al. 2013). Santucci et al. (2017) have driven this approach to the extreme case because they proposed building a classifier ensemble based on one unique model by varying its parameters.

Wozniak et al. (2014) proposed a survey on multiple classifier systems and suggested these systems for various applications, including remote sensing, computer security, banking, medicine, and recommender systems. Although in their study, Zhou et al. (2013) used an ensemble of surrogates for dual response surface modelling, which is a regression problem, the application of classifier ensembles to production control problems and particularly to the quality monitoring problems has not been investigated.

Moreover, some authors used different types of classifiers in their applications (Aksela and Laaksonen 2006, Yang 2011); however, to the best of our knowledge, there is no study on the impact of using different types of classifiers on model accuracy.

Therefore, to bridge these abovementioned gaps in research, we propose a methodological approach to build a classifier ensemble based on four types of classifiers, namely decision tree (DT), kNNs, multilayer perceptron (MLP), and SVM classifiers; in addition, we analyze the accuracy of proposed models on a real-world quality monitoring problem. In our study, we focus on the impact of the choice of classifiers types, the selection criterion, and the fusion process of the classifier ensembles. The best classifier (individual or ensemble) was selected to be implemented for the real application. Furthermore, the use of this classifier to determine the optimal setting of the controllable parameters is also presented.

The rest of the paper is organized as follows: Section 2 presents the related state-of-the-art methods and methodology for the classifiers used in this study and the construction of a committee of classifiers in general. The use of these classifiers to predict, and consequently, limit the incidence of quality defects is discussed in Section 3. Section 4 presents information about the application of our proposed classifiers for quality monitoring of a robotic lacquering workstation, wherein we compared the diversity and accuracy of the different techniques. In addition, some strategies for designing classifier ensembles were compared, and the optimal parameters for online quality were also investigated. Then, the use of classifiers to predict the incidence of defects and to determine the optimal setting of controllable parameters to limit defects incidence is illustrated on a real-world industrial problem. Although in this industrial application, 25 different types of defects may occur, our study focuses on only one of them. Finally, Section 5 presents the discussion and conclusion for this paper.

# 2 Related Work

The goal of classifier algorithms is to use a dataset to develop a model that classifies different instances into appropriate classes (Kotsiantis 2007). Köksal et al. (2011) presented a review of data mining applications for quality improvement. They classified these applications into four primary domains:

- Product and Process Quality Description (identifying and ranking the quality variables) (Hsu and Chien 2007, Han et al. 1999, Li et al. 2016),
- Quality Prediction, when quality may be represented with a real variable (Chen et al. 2007, Yang et al. 2005),
- Quality Classification, when the quality characteristics are binary nominal or ordinal (Krimpenis et al. 2006, De Abajo et al. 2004),
- Parameter Optimization (Hsieh and Tong 2001, Hung 2007).

The problem considered in this study is related to the last two categories mentioned above. The concept used here in designing a forecasting system is to ensure that it is in line with a physical system (Figure 1). In the case considered in this study, the forecasting model must predict the class (defect or no defect) as the output based on the parameter values collected from the real-world system. In addition, this forecasting model may be subsequently used to evaluate the decisions taken.

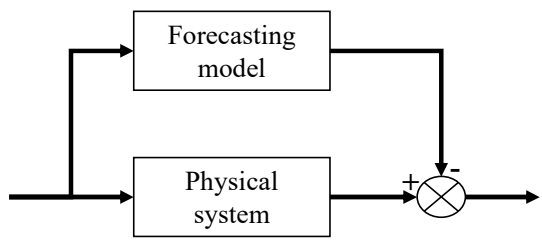

Figure 1: Forecasting system in parallel with the physical system.

The design of the forecasting system is based on knowledge discovery and data mining composed of a dataset collection task (variables selection, data collection, and preprocessing) and data mining task (classifier type choice, classification of the dataset into learning and validation datasets, fitting of the model, and evaluation of the results). To improve accuracy of the classifier, a classifier ensemble may be used.

## 2.1. Methodology for designing the classifier ensemble

In the past decade, the classifier ensemble has been established as a significant research field (Wozniak et al. 2014). The goal of the classifier ensemble is to combine a collection of individual classifiers that are not only diverse but also accurate. In particular, considerably accurate classification methods can be developed by combing the decisions of the individual classifiers in the ensemble based on some voting method (Dietterich 2000). A classifier ensemble can be built at four different levels (Kuncheva 2004), namely the data (Breiman 1996), feature (Ho, 1998), classifier, and combination levels (Kuncheva 2002). The design of a classifier ensemble consists of two primary steps: the generation of multiple classifiers and then, their fusion (Dai 2013); this leads to two challenging issues; first, how to select the individual classifiers, and second, how to combine the selected classifiers.

The key to designing a successful ensemble is to ensure that the classifiers in the group are sufficiently diverse. The four most popular algorithms (Kotsiantis 2011) for improving diversity are the bagging (Breiman 1996), boosting (Freund and Schapire 1996), rotation forest (Rodriguez *et al*. 2006), and random subspace methods (Ho 1998). The bagging and random subspace methods are more robust than other algorithms when the data are noisy (Dietterich 2000, Kotsiantis 2011); therefore, for this study, we selected the bagging approach.

### 2.1.1. Selection criteria for individual classifiers

Different selection criteria have been proposed in the literature. The individual performance is a universal indicator for selecting the best individual classifier. It is the simplest method, and is reliable and robust. Therefore, it is generally preferred in industrial applications (Ruta and Gabrys 2005). The minimum individual error (MIE) indicator represents the minimum error rate of the individual classifier and is used in the selection strategy for the best individual classifiers. MIE is mathematically defined as:

$$MIE = \min_j \left( \frac{1}{n} \sum_{i=1}^{n} e_j(i) \right), \qquad (1)$$

where $e_j(i)$ represents the classification error of classifier $j$ for data $i$.

Another approach for classifier selection is to consider the diversity of the individual classifiers; however, there is no consensus about the definition of a measurement metric for diversity. Consequently, there are different diversity measures in the literature, such as the Q statistics, correlation, disagreement, double fault, entropy of votes, difficulty index, Kohavi-Wolpert variance, inter-rater agreement, and generalized diversity; many authors have tested and compared these measures using different examples (Kuncheva and Whitaker 2003, Tang *et al*. 2006, Aksela and Laaksonen 2006, Bi 2012). Kuncheva and Whitaker (2003) showed that all these measures yield similar results; therefore, in our study, we use the double-fault (DF) measure to evaluate the diversity between the different classifiers. The DF metric was proposed by Giacinto and Roli (2001) to form a pairwise diversity matrix for a classifier pool; it selects the classifiers that are least related. Further, this measure is defined by the proportion of cases that have been misclassified by both classifiers considered, i.e.,

$$DF_{i,j} = \frac{N^{00}}{N^{11} + N^{10} + N^{01} + N^{00}}, \qquad (2)$$

where $i$ and $j$ represent the two classifiers, and $N^{ab}$ is the number of simultaneous correct outputs of classifier $i$ when $a = 1$ (or incorrect, if $a = 0$) and correct outputs of classifier $j$ when $b = 1$ (or incorrect, if $b = 0$). For example, $N^{00}$ represents the number of outputs for which, both classifiers are incorrect. Thus, the lower this value is, the greater the diversity between the two considered classifiers.

### 2.1.2. Selection process for individual classifiers

The classifier selection problem has been addressed by many authors in previous studies (Ruta and Gabris 2005, Hernandez-Lobato 2013, Dai 2013); different strategies have been proposed to select the classifiers for an ensemble (Kim and Oh 2008, Ko et al. 2008, Tsoumakas et al. 2009, Yang 2011, Guo and Boukir 2013, Soto et al. 2013).

In this study, two strategies of classifier selection are tested and compared to evaluate the impact of the diversity of classifiers on the accuracy of the classifier ensemble.

The first strategy is based on the accuracy of individual classifiers as given by Eq. (1); based on this, the classifiers are added to the ensemble sequentially from the more accurate to the less accurate one, and the selected structure is one that yields the best classification result on a given validation dataset.

The second strategy is based on the accuracy and diversity (SAD) approach proposed by Yang (2011). It is a simple recursive strategy consisting of the following steps:

1. Train a set of different classifiers.
2. Evaluate each classifier's accuracy using the validation dataset.
3. Select the most accurate classifiers to perform a vote.
4. Calculate the diversity between the classifier ensemble and the remaining individual classifiers using the double-fault measure.
5. Select the classifier with strong diversity and add it to the classifier ensemble to construct a new ensemble.
6. Evaluate the performance of the new classifier ensemble. If all the classifiers are not used, repeat Steps 4–6; otherwise, compare all the classifier ensembles and select the best ensemble.

### 2.1.3. Fuser design

The classifiers have been classically fused using a majority vote. However, Kuncheva *et al.* (2003) showed that the use of a majority vote for fusing classifiers may lead to degradation of accuracy; in particular, they highlighted the lack of independence in the classifiers, which may limit the benefits of classifier ensembles, because the majority vote is performed on the binary outputs of the individual classifiers.

It is noteworthy that some types of classifiers, such as MLP or SVM, yield a real value as output. This real value can be seen as a membership function, which must be compared to a threshold in order to determine the class of the considered pattern; for example, with a classical threshold of 0.5, an output value of 0.49 corresponds to the association of the considered pattern to class 0, whereas a value of 0.51 associates the pattern to class 1. However, a disadvantage of this approach is the high risk of losing some information; consequently, to address this issue in the case of MLP and SVM classifiers, we propose a second fusion approach, which involves calculating a mean of these real outputs, which can be compared to the threshold as a second step to associate the pattern to an appropriate class.

Another approach to perform the fusion of classifiers is to consider this problem as a specific learning process, for example, Wozniak trained the fuser using perceptron-like learning (Wozniak, 2007) and an evolutionary algorithm (Wozniak 2009). In this study, an MLP approach, which is described in subsection 2.2.3, will be used to perform the fusion process; this process is shown in Figure 2. The outputs of the individual classifiers are used as inputs of the MLP used in our study; then, the learning step allows the fusion of the outputs of the individual classifiers. Finally, the selection of the classifiers is performed via pruning.

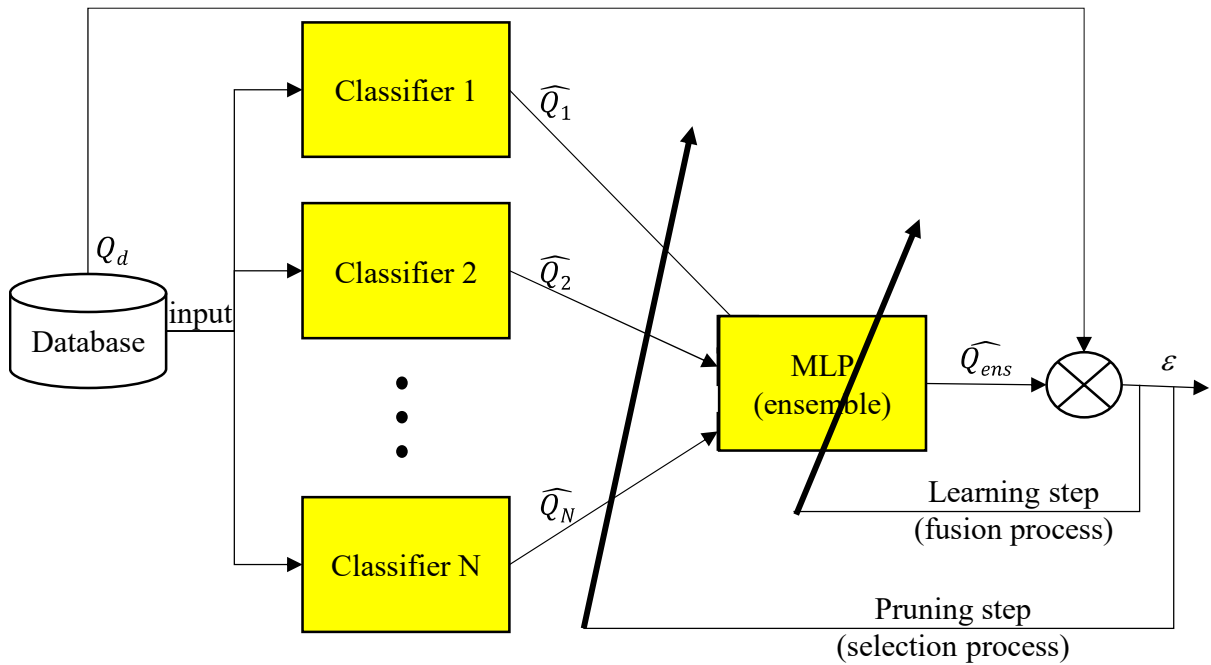

Figure 2: NN fusion process.

These three approaches are referred to as vote, mean, and neural network (NN) fusion processes, respectively; these approaches were tested and compared based on two selection criteria, namely accuracy and diversity.

## 2.2.    Selected individual classifiers

Our study focuses on supervised problems with continuous inputs wherein four approaches may be exploited to perform data mining (Jain et al. 1999); these include the logic-based, neural network, SVM, and instance approaches. In order to evaluate the impact of the different types of classifiers on the accuracy of the classifier ensemble, one classifier from each type of approach is considered.

### 2.2.1. Decision trees

Decision trees are sequential models that logically combine a sequence of simple tests. Each node in a decision tree represents a feature in an instance to be classified, and each branch represents a value that the node can assume. Instances are classified starting at the root node, and sorted based on their feature values (Kotsiantis 2013).

Numerous decision tree algorithms have been developed, including the C4.5 (Quinlan 1996), classification and regression tree (CART) (Breiman et al. 1984), SPRINT (Shafer et al. 1996), and SLIQ (Mehta et al. 1996). In this paper, the CART decision tree algorithm proposed by Breiman *et al.* (1984) with the Gini impurity function (Breiman et al. 1984, Mathworks 2007) is selected.

### 2.2.2. kNN

kNN is a non-parametric lazy learning algorithm, which uses databases; in this algorithm, data are separated into several classes to predict the classification of new samples (Cover and Hart 1967). The classification of new samples is performed by determining the label of an unlabeled example by a majority vote of its k nearest neighbors in the training set. The accuracy of the kNN classification depends on the metric used to compute the distance between the examples. Different metric distances have been proposed to determine the neighbors; however, in this study, the Mahalanobis distance is used, which considers non-scattered inputs as more important than scattered ones.

Furthermore, the choice of k is critical, because the smaller the value of k, more is the effect of noise on the classification; however, a higher value of k leads to a significant increase in computational costs and also leads to the loss of model flexibility (James et al. 2013). In the implementation of kNN used in this study (Ballabio and Consonni 2013), the selection of k is performed by cross-validation.

### 2.2.3. MLP

Artificial NNs have been successfully applied to solve various problems, including dynamic system identification, pattern classification, adaptive control, and the approximation function. Among the types of NN, MLP is, by far, the most popular (Han and Qiao 2013); in MLP, the hyperbolic tangent is used for hidden neurons, whereas a sigmoid is used for the output neuron. The NN model requires three design steps:

- **Initialization**: This step involves the determination of the initial set of weights and biases; it has an impact on the local minimum trapping problem and diversity of the NN classifiers. In this study, the Nguyen Widrow algorithm is used (Nguyen and Widrow 1990, Mathworks 2016).

- **Training**: The training algorithm fits the network output to the data. Data from industrial applications are noisy and corrupted. To limit the impact of outliers on the results, a robust Levenberg-Marquardt algorithm is used here (Thomas *et al*. 1999, Mathworks 2016).

- **Pruning**: This step involves determination of the optimal structure of the network, and is crucial to avoid the overfitting problem. This step is important for ensembles of NNs, because it simultaneously improves the performances of each classifier and their diversity by giving them different structures. The pruning procedure used is the one proposed by Thomas and Suhner (2015) (Mathworks 2016).

### 2.2.4. SVM

SVMs and NNs are based on considerably similar concepts and therefore yield similar results. Related research shows that, in some cases, the SVM method produces better results than the NN method (Meyer *et al*., 2003), whereas, in other cases, the NN method performs better (Paliwal and Kumar, 2009).

In particular, SVMs are an effective approach for the classification and nonlinear function estimation problems (Cortes and Vapnik 1995, Platt 1998, Yang et al. 2010). Further, SVM solutions are characterized as convex optimization problems. Theoretically, SVMs have been distinctive from other machine learning algorithms and NNs because they provide a global and unique solution rather than multiple local minima (Christianini and Shawe-Taylor 2000). However, these optimization algorithms may provide different numbers of support vectors (Fan et al. 2005, He et al. 2012, Liang et al. 2013).

SVMs can be considered as an extension of the linear learning machines that linearly divides the input space. However, complex real-world applications require more expressive hypothesis spaces than linear functions. Kernel representations offer a solution by projecting the data into a high-dimensional feature space, increasing the computational power of the linear learning machines (Christianini and Shawe-Taylor 2000). Different kernels have been proposed in the past, such as linear, polynomial, radial basis function (RBF), and sigmoid kernels. Hsu *et al*. (2003) proposed the use of an RBF kernel, which uses a lesser number of hyper-parameters than a polynomial kernel. The learning algorithm used in this study is the one proposed by Grandvalet and Canu (2008).

In addition, SVMs share many advantages with MLPs; SVMs extract models from the data and adapt them to different changes of the system or its environment, using re-training strategies (Noyel et al. 2016). Moreover, because there is no local optimum in the SVM algorithm, considerably stable models are obtained that are less sensitive to noise or small variations in the modeled system. Although the stability of the resulting model is an advantage when we are only constructing one classifier, it becomes a drawback in terms of model diversity. In addition, this stability can limit the impact of bagging strategies on diversity.

# 3 Proposed Approach based on Classification Model for Online Parameter Tuning

In the case of a quality monitoring problem, a classifier or ensemble model could predict the risk of quality defect occurrence as a function of the operating point. Therefore, it is possible to use such a classifier ensemble model before processing the product on a workstation to prevent the risk of defects.

The concept of our proposed approach involves exploiting the prediction capabilities of the classifier to determine the optimal setting of the controllable factors (Figure 3). The impacting/critical factors are divided into controllable (parameters, which may be tunable) and uncontrollable factors (environmental factors, or factors restricted by the production range). The set of uncontrollable factors correspond to the operating point. Each production batch corresponds to one operating point, which may need a particular setting of the controllable factors. Therefore, an experimental design for the classifier model is required for simulation to determine the optimal tuning of the controllable factors for the considered operating point (Figure 3).

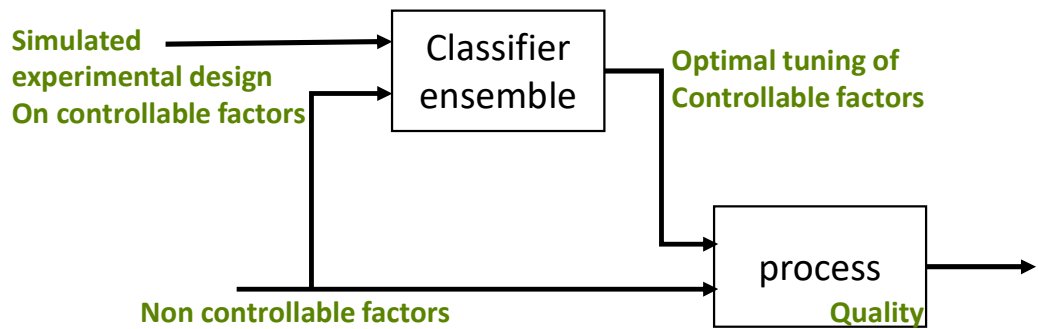

Figure 3: Proposed approach to obtain the optimal setting of the controllable factors.

# 4 Industrial Application: Quality Monitoring Problem

### 4.1. Process

The industrial partner company we collaborated with for this study produces high-quality lacquered panels made with medium density fiberboard (MDF) for kitchens, bathrooms, offices, stands, shops, and hotel furniture. One of its critical processes involves a robotic lacquering workstation. Even though this workstation is free of human factors, the quality of the products manufactured is unpredictable (because we cannot know if there is a risk that the product has one or more defects) and fluctuant (the percentage of defects may be variable at different times even with the same settings). In our study, we try to improve the process quality by focusing on the quality of the product. In particular, the aim is to find the optimal setup of the robotic workstation in a changing environment that can impact the quality of production. The classical quality management tools (Statistical Process Control (Oakland 2007), Taguchi Optimal Experimental Design (Taguchi 1989), etc.) are not sufficient to control and monitor the product quality. Consequently, we propose developing a prediction model for quality using a classifier ensemble that is able to link the impacting factors, including product characteristics, environmental conditions, and process parameters, to the incidence of quality defects. There are many possible defects that can occur on lacquer, some of which are well-known in the industrial best practices guidelines (Standox 2018); for example, dripping, bubbling, stains, and low opacity. Some defects can be detected directly after drying (dripping, bubbling, etc.), but others can only be identified after the following operations like polishing are completed, such as, the lack of material at the product edge that causes a loss of color during polishing referred to as a "perce." For our process, we worked on the 25 most common quality defects for which operators are able to detect and notify about the defect at the drying exit stage or after. The company categorizes these 25 types of defects as a function of the extra costs they incur. Table 1 presents the repartition of these different types of defects as a function of their extra costs ( scrap, rework, raw materials….)

| extra cost | low | medium | expansive |
|---|---|---|---|
| nb of defect's types | 7 | 17 | 1 |

Table 1: Repartition of defect types as a function of their extra costs.

We built a forecasting model for each of the 25 defect types, consequently, a complete quality monitoring system that is able to predict the existence of all 25 defect types is obtained. These forecasting models may be extracted from the dataset using a data mining approach (Yu et al. 2008, Xiaoqiao et al. 2015). For simplification of the presentation of our approach, in this study, we focus on only one of these 25 defect types, namely, "stain on back," for two reasons: the associated cost is considered medium, and it has the highest occurrence frequency.

The factors that influence the quality of this process were identified based on subject-matter expert knowledge, using some quality tool like the Ishikawa diagrams (Ishikawa 1986). The production and quality management system recorded the value of these factors for eight months, which were used to build the dataset. We use the Manufacturing Execution System (MES) to collect and share these data. First, human experts highlight 11 critical factors impacting the "stain on back" defect whose values need to be observed, including the load factor (f8), number of passes (f4), time per table (lacquering batches) (f1), liter per table (f2), basis weight (f6), number of layers (f5), number of products (f3), and drying time (f7). In addition, environmental factors were automatically recorded, including temperature (f9), atmospheric pressure (f10), and humidity (f11). Then, using an experimental design approach, these factors were classified into two types: internal (load factor, number of passes, time per table, liter per table, basis weight, number of layers, number of products, and drying time) and external (temperature, humidity, pressure) (Figure 4). Only factors f4 and f5 are discrete and can attain one of the three states, whereas all the others 9 factors are continuous. Consequently, for each classifier (NN, Tree, kNN, or SVM), there are 15 inputs (9 continuous factors and two discrete factors with three states).

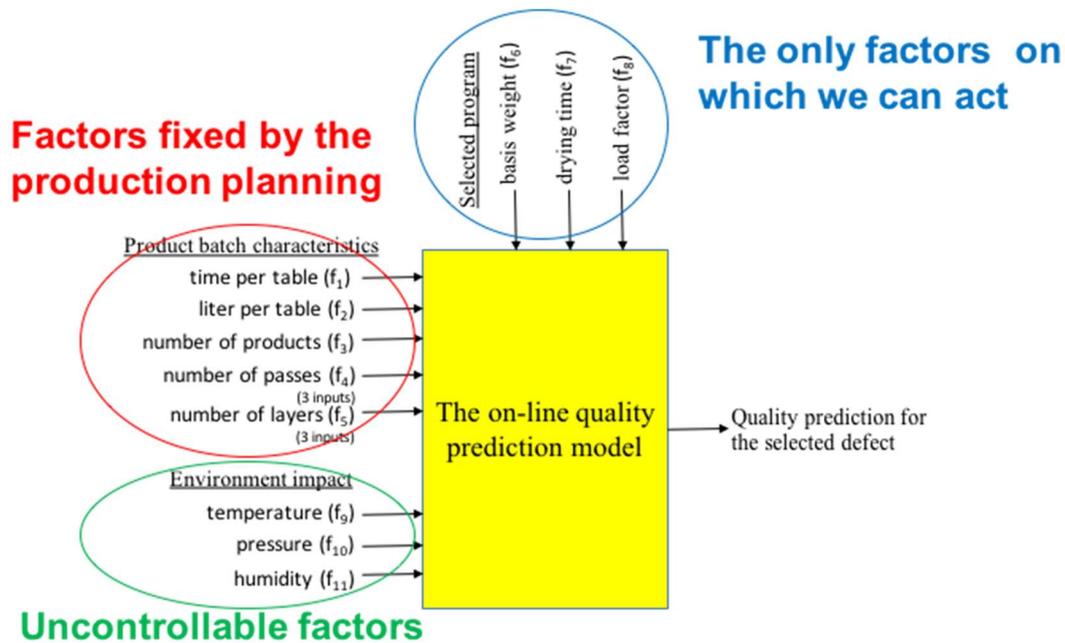

Figure 4: Online quality prediction model using the classifier ensemble linking the impacting factors to the considered type of defect.

## 4.2. Experiment

### 4.2.1. Construction of classifiers

For the classification model, a sufficiently large dataset (2270 items) is required to use the holdout validation method. Therefore, our dataset is randomly divided into two datasets: one for identification (1202 items) and the other for model validation (1068 items). A bagging strategy is applied to the training dataset to improve the diversity of the individual classifiers; however, this step is not necessary for NNs, because the results obtained in the latter case are equivalent with or without bagging. For the experiment, the procedures described in Section 2.2 are used to construct 100 different tree classifiers, 100 different NN classifiers, 100 different kNN classifiers, and 100 different SVM classifiers. It is important to note that the 100 different NN classifiers used in this study were the same ones used in a previous work (Noyel *et al.* 2013). All the algorithms were implemented in MATLAB® 2015 and executed on a PC with an Intel i5 processor, 8 GB of RAM, running the Windows® 7 Pro operating system.

Then, the diversity of the different models is investigated using Eq. (2); in addition, the accuracy of the individual classifiers was compared. Finally, 19 classifier ensembles were constructed and compared, including those using only NN models, only Tree models, only kNN models, only SVM models, and all four types of models used simultaneously (NN, SVM, kNN, and Tree) with selection processes based on diversity or accuracy of the classifiers. The fusion of the classifiers is performed using a vote strategy, mean strategy, or by building a NN.

### 4.2.2. Comparison criterion for the classifiers

In classification problems, the goal is to reduce the number of misclassified data. Thus, the classical criterion for classification problems is the misclassification rate (error rate or "zero-one" score function) (Hand et al. 2001):

$$S_{01} = \frac{1}{N} \sum_{n=1}^{N} I(y_n, \hat{y}_n) \quad (3)$$

where $I(a, b)$=1 when $a \neq b$ and 0 otherwise, and $N$ is the number of patterns.

Another difficulty arises if there are different costs associated with different misclassifications; in this case, a general loss matrix must be constructed (Bishop, 1995). However, this problem is not considered here; instead, the best misclassification rate $S_{01_{min}}$ was obtained on the validation dataset using the best approach. Then, a McNemar statistical hypothesis test is used to determine whether the misclassification rate of other approaches is statistically different from that obtained with the best one. The null hypothesis (the tested approach is statistically equal to the best one) was tested, where $\mathcal{H}_0$ and its alternative $\mathcal{H}_1$ are defined as follows:

$$\begin{cases} \mathcal{H}_0 : \quad S_{01} = S_{01_{min}} \\ \mathcal{H}_1 : \quad S_{01} \neq S_{01_{min}} \end{cases} \quad (4)$$

The null hypothesis $\mathcal{H}_0$ is rejected with a risk of 5% if:

$$U = \frac{|N_{10} - N_{01}|}{\sqrt{N_{10} + N_{01}}} > 1.96 \quad (5)$$

where $N_{10}$ is the number of cases where the best classifier correctly classifies a data item, whereas the compared classifier does not, and $N_{01}$ is the number of cases where the best classifier misclassifies a data item, whereas the compared classifier yields the correct class. Two other criterions are also considered, the false alarm rate (FA) and the non-detection rate (ND) given by:

$$\begin{cases} FA = \frac{FP}{FP+TN} \\ ND = \frac{FN}{FN+TP} \end{cases} \quad (6)$$

where *FP* is the number of false positives, *TN* is the number of true negatives, *FN* is the number of false negatives, and *TP* is the number of true positives.

### 4.3. Results

#### 4.3.1. Diversity between individual classifiers

We studied the diversity between the different classifiers. In that light, we used the DF diversity measure given by Eq. (2), which is a pairwise measure and yields a lower value for greater diversity. In particular, this measure indicates the percentiles of data for which, the two considered classifiers provide incorrect results.

Figure 5 presents the distribution of the DF measure for the four classifiers individually and all of them together; in the figure, the box represents the second and third quartiles of the distribution. It is clear from the figure that even if the SVM and kNN approaches lead to different models, these models are less diverse than those obtained using the other approaches. In general, the tree approach leads to better diversity between models even though the smallest value of the DF measure is obtained with the NN models, i.e., the NN approach leads to more scattered DF values.

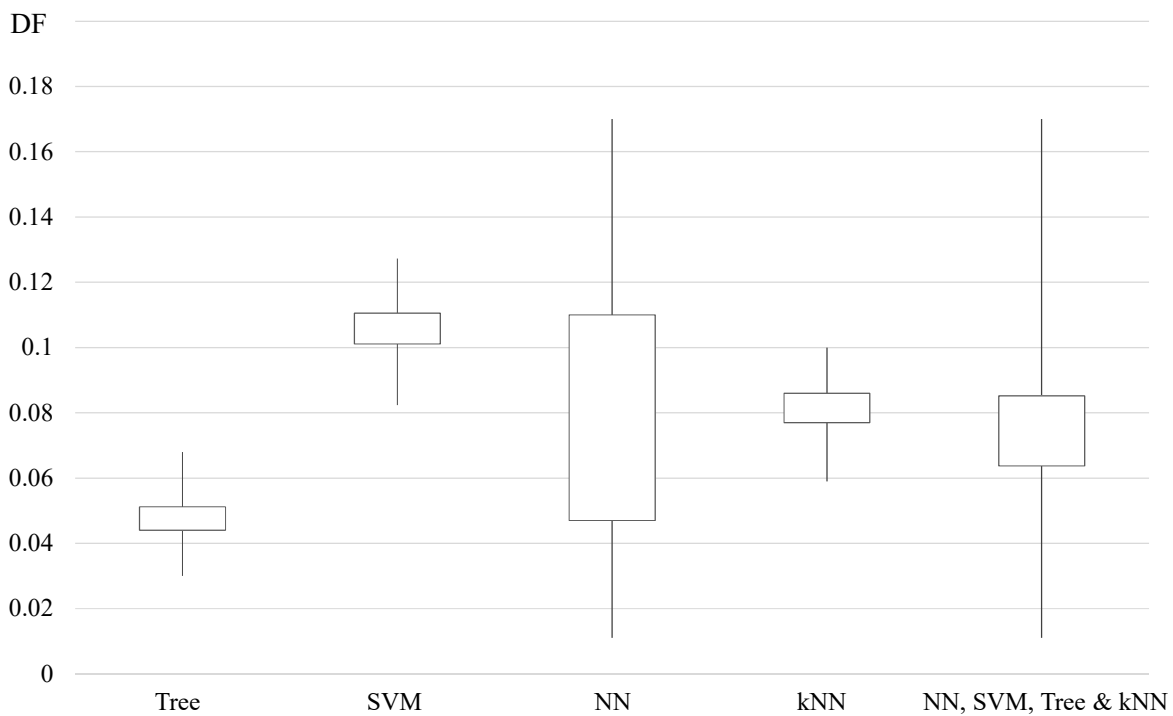

Figure 5: Distribution of the DF diversity measure for different classifiers.

### 4.3.2. Accuracy of individual classifiers

The diversity between the different classifiers is important; however, it is less important than their accuracy. Figure 6 shows the misclassification rates for the different classifiers; the box represents the second and third quartiles of the distribution.

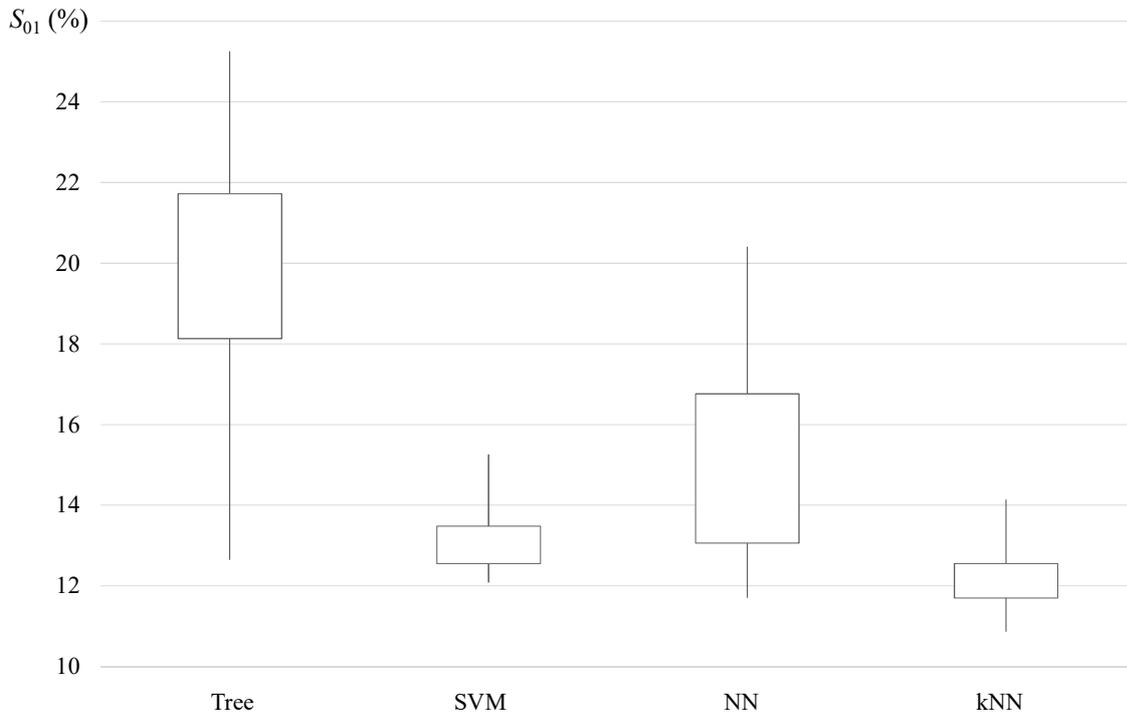

Figure 6: Distribution of the misclassification rates for different classifiers.

This figure shows that, in general, the kNN classifier yields the best results, whereas the Tree classifier is the weakest. Furthermore, different SVM models lead to different results, but with small variability, whereas the NN models yield more scattered results.

Table 2 shows the best results obtained with the different individual classifiers. Although the kNN models give the best results for the error rate, they have a poor non-detection rate, similar to the SVM models. In contrast, the NN and Tree models yield similar results with a better non-detection rate, but with a higher false alarm rate.

|  | error rate | false alarm rate | non-detection rate | U |
|---|---|---|---|---|
| best NN classifier | 11.7% | 9.9% | 25.2% | 3.91 |
| best Tree classifier | 12.6% | 6.9% | 55.1% | 5.12 |
| best knn classifier | 10.9% | 1.9% | 77.2% | 3.40 |
| best SVM classifier | 12.1% | 1.9% | 87.4% | 4.54 |

Table 2: Results obtained for the best NN, Tree, kNN, and SVM classifiers.

Table 3 presents the confusion matrix for the two best individual classifiers (NN and kNN). The information in this table indicates that these two classifiers have different behaviors. In particular, the kNN models have lower false alarm rates, whereas the NN model is more balanced leading to a good non-detection rate.

| | best NN classifier | | | best knn classifier | | |
|---|---|---|---|---|---|---|
| | Defect (predicted) | non defect (predicted) | accuracy | Defect (predicted) | non defect (predicted) | accuracy |
| Defect | 95 | 32 | 74.80% | 29 | 98 | 22.83% |
| Non defect | 93 | 848 | 90.12% | 18 | 923 | 98.09% |

Table 3: Confusion matrix of the two best individual classifiers (NN and kNN).

However, the results obtained with these four types of models are considerably similar and the holdout validation method is not sufficient to compare them. Various other validation approaches (Arlot and Celisse 2010) may be used for this comparison, such as the bootstrap (Efron and Tibshirani 1997), k-fold cross validation (Kohavi 1995), or leave-one-out validation, which can be considered as a particular case of the k-fold cross validation method (Efron *et al.* 1998). Efron et al. (1998) showed that the bootstrap method is biased with small variance, whereas the k-fold cross validation method is unbiased, but with large variance. Consequently, a stratified k-fold cross validation (Kohavi 1995) is used here, where k is classically fixed to 10. Table 4 shows the results of the 10-fold cross validation for the four tools. The mean and the standard deviation of the error rate, false alarm rate, and non-detection rate obtained on the 10 folds are presented; these results show that the NN classifier yields the best results with a small variance. In addition, the SVM and kNN classifiers give similar results for the error rate, but with a significantly worse detection rate in the case of the SVM approach. However, the Tree approach yields the worst results with the worst variance. Table 4 lists the computational time (mean and standard deviation) obtained for learning in the case of different models during the k-fold cross validation. As can be seen from Table 4, the Tree algorithm is the quickest approach, whereas the SVM is the slowest. It should be noted that the computational time for the NN classifier excludes the pruning step, which is computationally demanding.

| | error rate | | False alarm rate | | Non-detection rate | | Computational time (s) | |
|---|---|---|---|---|---|---|---|---|
| | Mean | STD | Mean | STD | Mean | STD | Mean | STD |
| NN classifier | 10.9% | 2.7% | 7.3% | 1.9% | 37.5% | 12.9% | 2.68 | 0.31 |
| Tree classifier | 12.5% | 4.5% | 7.0% | 3.2% | 52.8% | 17.6% | 0.19 | 0.01 |
| knn classifier | 12.1% | 2.4% | 4.5% | 1.8% | 68.1% | 7.2% | 7.02 | 0.57 |
| SVM classifier | 12.4% | 2.0% | 1.4% | 0.8% | 93.0% | 5.3% | 11.20 | 0.73 |

Table 4: Results obtained for the 10-fold cross validation for individual classifiers.

### 4.3.3. Design of classifier ensembles

Based on the abovementioned results, though the best kNN classifier performs well with respect to misclassification rates and the best NN and Tree classifiers have good non-detection rates, these results are not sufficiently good and must be improved further; therefore, classifier ensembles should be used. Table 5 lists the misclassification, false alarm, and non-detection rates for the best classifier ensembles wherein the individual classifiers were selected according to the accuracy or diversity metrics and the fusion was performed using vote, mean, or NN strategies.

| | Selected based on | Fusion method | Size | error rate | False alarm rate | Non-detection rate | U |
|---|---|---|---|---|---|---|---|
| NN ensemble | | vote | 12 | 8.8% | 5.2% | 35.4% | 1.39 |
| NN ensemble | | mean | 13 | 8.6% | 6.4% | 25.2% | 1.13 |
| Tree ensemble | | vote | 12 | 10.3% | 2.1% | 70.9% | 2.95 |
| kNN ensemble | Accuracy | vote | 23 | 10.3% | 0.5% | 82.7% | 2.77 |
| SVM ensemble | | vote | 1 | 12.1% | 1.9% | 87.4% | 4.54 |
| SVM ensemble | | mean | 9 | 11.4% | 2.0% | 81.1% | 4.00 |
| Classifier ensemble | | vote | 6 | 9.0% | 3.1% | 52.8% | 1.65 |
| NN ensemble | | vote | 17 | 10.7% | 8.1% | 29.9% | 3.57 |
| NN ensemble | | mean | 22 | 9.0% | 5.4% | 35.4% | 1.94 |
| Tree ensemble | | vote | 10 | 10.1% | 1.4% | 74.8% | 2.71 |
| kNN ensemble | Diversity | vote | 28 | 10.2% | 1.6% | 74.0% | 2.71 |
| SVM ensemble | | vote | 23 | 11.7% | 1.6% | 86.6% | 4.37 |
| SVM ensemble | | mean | 1 | 12.1% | 1.9% | 87.4% | 4.54 |
| Classifier ensemble | | vote | 24 | 7.7% | 4.0% | 34.7% | - |
| NN ensemble | | | 92 | 10.6% | 6.2% | 43.3% | 2.94 |
| Tree ensemble | | | 97 | 13.1% | 6.4% | 62.2% | 5.91 |
| kNN ensemble | NN | | 100 | 13.1% | 4.5% | 77.2% | 5.88 |
| SVM ensemble | | | 98 | 12.1% | 2.6% | 82.7% | 4.27 |
| Classifier ensemble | | | 394 | 12.0% | 1.1% | 92.9% | 4.20 |

Table 5: Misclassification, false alarm, and non-detection rates obtained using the best classifier ensembles.

The best result ($S_{01_{min}} = 7.7\%$) is obtained with the classifier ensemble using the four types of individual classifiers selected based on diversity. The McNemar test is performed to determine if the other approaches are statistically different compared with the best one; the test showed that the four best individual classifiers presented in Table 2 do not perform as well as the best classifier ensemble (U>1.96). In summary, the use of a classifier ensemble significantly improves the accuracy of classification.

The use of the NN strategy to fuse the classifiers does not yield a suitable structure for an efficient classifier; this is true even for the cases of the SVM, Tree, and kNN ensembles as well (U>1.96). However, this was expected for kNN and SVM ensembles owing to the weak diversity between the individual classifiers in these cases. In contrast, the Tree ensemble does not yield a suitable structure because of the weak accuracy of the individual classifiers. Four ensembles classifiers yield results that are statistically equivalent to the best one (U<1.96), including the two NN ensembles with selections based on accuracy, the NN ensemble with selections based on diversity with the mean fusion approach (considerably close to the threshold of 1.96), and the classifier ensemble with selections based on accuracy.

The mean fusion approach slightly improves the results for NN ensemble (with selection based on accuracy or diversity). The best result is obtained with the classifier ensemble with selection based on diversity, however the classifier ensemble with selection based on accuracy is the more parsimonious (only 6 individual classifiers).

Another advantage of the classifier ensemble is that the vote used for fusion may be used as a confidence interval on the classification. For example, if 40% of the classifiers votes for a defect and 60% votes for a non-defect, we may deduce that there is a relatively high chance of a defect.

In conclusion, five classifier ensembles lead to results that sufficiently close such that the holdout validation method cannot discriminate between them leading to five statistically equivalent ensembles (U<1.96). Furthermore, to compare the four best individual classifiers, a stratified k-fold cross ($k$=10) validation is performed. Table 6 lists the results obtained for the 10-fold cross validation for the abovementioned five statistically equivalent classifier ensembles; these results indicate that the NN ensembles (with selection based on accuracy or diversity with the two fusion processes) tend to favor the non-detection rate (mean and standard deviation) compared with the classifier ensemble (with selection based on accuracy or diversity). However, the classifier ensemble with selection based on diversity leads to the best mean value of $S_{01}$. Based on these results, the following ensembles are the best two:

- Classifier ensemble with selection based on diversity and the vote fusion approach, which favors the global misclassification rate.

- NN ensemble with selection based on accuracy and the mean fusion approach, which favors the non-detection rate.

| | Selected based on | Fusion method | error rate | | False alarm rate | | Non-detection rate | |
|---|---|---|---|---|---|---|---|---|
| | | | Mean | STD | Mean | STD | Mean | STD |
| NN ensemble | Accuracy | vote | 5.2% | 1.5% | 1.0% | 0.7% | 18.8% | 9.9% |
| | | mean | 4.7% | 1.8% | 2.9% | 1.3% | 19.0% | 9.7% |
| classifier ensemble | | vote | 4.5% | 1.7% | 0.2% | 0.2% | 29.0% | 11.9% |
| NN ensemble | Diversity | mean | 5.6% | 1.4% | 0.9% | 0.6% | 18.7% | 8.7% |
| classifier ensemble | | vote | 3.3% | 1.3% | 0.5% | 0.4% | 22.8% | 10.1% |

Table 6: Results obtained for the 10-fold cross validation for the best ensembles.

Table 7 presents the confusion matrix for the two best ensemble classifiers (classifier ensemble with selection based on diversity and the NN ensemble with selection based on the accuracy using the mean fusion approach); from the table, it can be seen that the NN ensemble classifiers tend to have a more well-balanced behavior considering the false alarm and non-detection rates, whereas the classifier ensemble slightly favor the false alarm rate. Compared with the results listed in Table 3, we can conclude that the use of NN ensemble improves both non-detection and false alarm rates. Thus, the classifier ensemble with selection based on diversity is selected to be implemented on the real-world system.

| | Classifier ensemble (diversity) | | | NN ensemble (accuracy and mean) | | |
|---|---|---|---|---|---|---|
| | Defect (predicted) | Non defect (predicted) | Accuracy | Defect (predicted) | Non defect (predicted) | Accuracy |
| Defect | 83 | 44 | 65.35% | 95 | 32 | 74.80% |
| Non defect | 38 | 903 | 95.96% | 60 | 881 | 93.62% |

Table 7: Confusion matrix for the two best classifier ensembles

### 4.3.4. Implementation of the forecasting system

The steps described above were also performed for the 24 other defect types leading to the design of 25 quality prediction models (classifier ensembles). All these forecasting models are embedded in the supervision tool of the lacquering workstation. The structure and the parameters of these models are remotely stored in an SQL database, which allows different independent software to access this information, if needed; in addition, these values can be updated remotely using the database. The participating company's philosophy requires maintaining humans in the loop. Therefore, the quality monitoring system in this study is designed as a decision support tool, and is included in the setup interface of the painting robot (Figure 7).

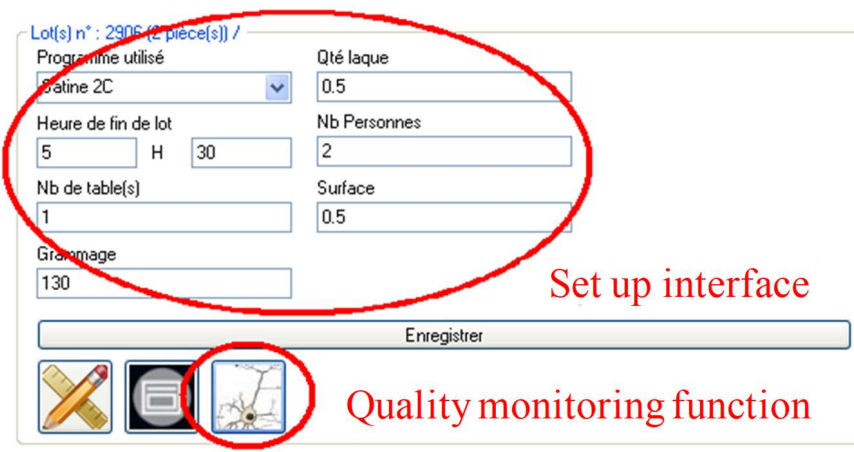

Figure 7: Setup interface of the painting robot.

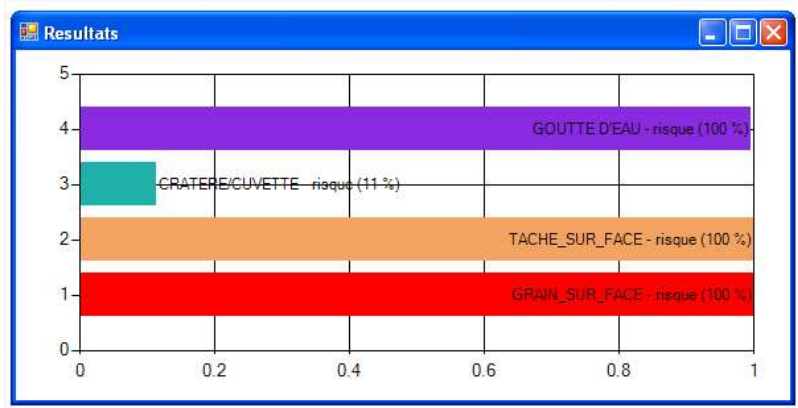

Figure 8: Forecast example.

This interface allows the operator to set up the robot's parameters (such as basis weight or production program); the lot characteristics and environmental conditions are input by the company's information system without operator intervention. After the information input, the quality monitoring function is available to the operator, if needed, and indicates the incidence risk of defect types' (Figure 8). If the operator judges that the risk is unacceptable, he can test another setting program, until the risk becomes acceptable. In reality, risk prediction is performed in approximately 12 seconds, allowing the operator to test different setting in hidden time.

### 4.3.5. Use of classifier ensembles for optimal online parameter tuning

For our research, an improved version of the abovementioned quality prediction system is also studied, even if the company does not wish to introduce it because it is contrary to their philosophy of keeping humans in the manufacturing loop.

As proposed in Section 3, the classifier ensemble is used instead of the real system to simulate experiments that achieve an entire DoE plan without cost. The aim is to find adequate/optimal values of the controllable factors that can help avoid or limit the risk of the incidence of defects for each operating point corresponding to the actual values of the non-controllable factors; however, this DoE plan cannot be achieved using the real system because it must be performed for each new configuration of uncontrollable factors (for example: for temperature changes, another DoE plan must be simulated; if a new product is considered with different production range, another DoE must be simulated). In fact, one DoE plan must be simulated for each new combination of environment conditions (temperature, pressure, etc.) and product range (time per table, liter per table, number of passes, etc.), leading to the simulation of one new DoE every two or three minutes.

As an example, Figure 9 shows the results of an entire DoE plan in which 10 levels were chosen for the 3 controllable factors for one operating point. In this example, the protocol factors corresponding to the batch characteristics that are tuned as follows:

- Number of passes and layers are set to 1;
- Time per table, and liters of lacquer are set to their average values;
- Number of products is set to its median value.

The environmental factors (temperature, humidity, pressure) are set to their average values.

The effect of each factor $x_i$ at level $A_i$ is classically obtained by calculating the mean of all the results (defects' incidence) obtained when $x_i = A_i$ subtracting the mean of the results obtained in case of all the experiences. When an effect is positive, it implies that the considered level increases the risk of defects' incidence, whereas when it is negative, the considered level reduces the incidence of defects. The interactions between factors may be investigated in a similar manner.

Figure 9 shows the results obtained with the best classifier ensemble using the diversity criterion.

Because of the diversity of the different classifiers, the impact of each effect is presented in the form of an envelope including the effects for all the classifiers, where the envelope is built by using the upper and lower values of the different classifiers for each level of the considered factor.

These results show that, for the considered operating point, some bounds can be found for the optimal tuning of the three controllable factors. For example, in order to limit the risk of the incidence of defects, the basis weight factor must be tuned between 96 and 135, while the dry time factor must be between 280 and 1660 minutes, and the load factor must be lower than 4. In comparison, the same analysis performed with a single classifier allows us to find only one of the two bounds for these controllable factors and thus, this is another advantage of using a classifier ensemble (Thomas *et al*. 2013).

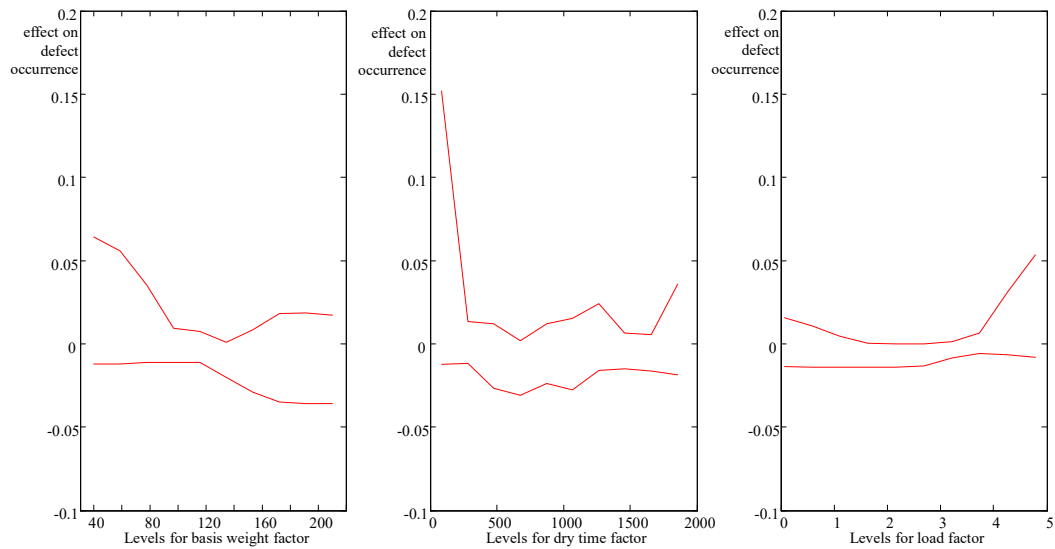

Figure 9: Experimental design results using the selected classifier ensemble.

In order to obtain the same results using a classical DoE, we would need to use 10 modalities for the three controllable factors, which would require many experiments, even when using a Taguchi plan. Moreover, this work would have to be performed for each batch and for each operating point, which is unrealistic.

# 5 Discussion and Conclusion

This study investigated the use of the NN, decision tree, kNN, and SVM classifiers as well as the benefits of using ensemble classifiers for a methodological approach to implement an online quality monitoring; this approach was applied and tested on an industrial application, specifically quality monitoring. The results of our experiment demonstrated that the use of classifier ensembles improves the accuracy of classification. Further, the combination of different types of classifiers (NN, SVM, kNN, and Tree) in the ensemble improves the results; however, using only NN type classifiers in the classifier ensemble leads to better results in terms of the non-detection rate.

The effect of the selection of individual classifiers based on either diversity or accuracy is still unclear. This is because, although a classifier ensemble with four types of classifiers with individual classifiers selected based on diversity always can yield the best results in terms of the misclassification rate, there is no statistical difference with a similar type of ensemble with selections based on accuracy. Moreover, in general, the selection based on accuracy leads to a more parsimonious ensemble.

Furthermore, although the SVM classifiers yield satisfactory results, the low diversity between the different SVM models prevents the improvement of the results using the classifier ensemble strategy; to a lesser extent, this is also true for the kNN classifiers.

The fusion of the individual NN classifiers in an ensemble using the mean strategy slightly improves the accuracy of the resulting ensemble; however, in general, there is no statistical difference between the two approaches.

In conclusion, the insignificant small improvement obtained by the combination of different types of classifiers is not sufficient to obtain a statistical difference compared with the use of only NN models for the creation of a classifier ensemble. The two strategies of selection based on accuracy or diversity have to be tested and compared, because both could yield good results with acceptable computational times. In addition, for similar reasons, the fusion processes using the mean or vote strategies have to be tested. However, using the NN approach to perform fusion of individual classifiers leads to an increased computational time with poor expected results

The results of this work provide many new and challenging perspectives in the case of both application and methodology. In the future, we aim to extend the use of the ensemble classifier to support the re-scheduling process, because, in the present study, the scheduling factors were unchanged. Moreover, owing to digital transformation adopted by many companies, there is continuous automatic collection of data leading to high data volume and short-term changes. Consequently, there is a real need to filter the useful data from the raw data. Therefore, we might explore two different alternatives for future work: (i) using an observer to determine if the classifier ensemble models need to be adapted or not, and (ii) designing scalable algorithms for classification; for example, the pruning method could be made more dynamic.

# 6 References

1. Aksela M., Laaksonen J., 2006. Using diversity of errors for selecting members of a committee classifier. Pattern Recogn., 39, 608–623.
2. Arlot S., Celisse A., 2010. A survey of cross-validation procedures for model selection. Statistics Surveys, 4, 40-79.
3. Ballabio D., Consonni V., 2013. Classification tools in chemistry. Part 1: Linear models. PLS-DA. Analytical Methods, 5, 3790-3798
4. Bi Y., 2012. The impact of diversity on the accuracy of evidential classifier ensembles. Int. J. Approx. Reason., 53, 584–607.

5.  Bishop C.M., 1995. Neural networks for pattern recognition. Clarendon Press, Oxford.

6.  Breiman L., 1996. Bagging predictors. Machine Learning, 24, 123–140.

7.  Breiman L., Friedman J.H., Olshen R.A., Stone C.J., 1984. Classification and regression trees, Chapmann & Hall, Boca Raton.

8.  Chen W.C., Lee A., Deng W.J., Liu K.Y. 2007. The implementation of neural network for semiconductor PECVD process. Expert Systems with Applications, 32, 1148-1153

9.  Christianini N., Shawe-Taylor J., 2000. An introduction to support vector machines and other kernel-based learning methods, Cambridge University Press, Cambridge, UK.

10. Cortes C. Vapnik V., 1995. Support vector networks. Machine Learning, 20, 273–297.

11. Cover T.; Hart P., 1967. Nearest neighbor pattern classification. IEEE Transactions in Information Theory, 21–27.

12. Dai Q., 2013. A competitive ensemble pruning approach based on cross-validation technique. Knowledge-Based Systems, 37, 394–414.

13. De Abajo N., Diez A., Lobato V., Cuesta S.R., 2004. ANN quality diagnostic models for packaging manufacturing: an industrial data mining case study. Proc. of the 10th ACM SIGKDD international Conference on Knowledge Discovery and Data Mining, Seattle, USA, 22-25 August

14. Dietterich T.G., 2000. An experimental comparison of three methods for constructing ensembles of decision trees: bagging, boosting, and randomization. Machine Learning, 40, 139–157.

15. Efron B., Tibshirani R., 1997. Improvement on cross-validation: the .632+ bootstrap method. J. of the American Statistical Association, 92, 548-560.

16. Efron B., Tibshirani R., 1998. An introduction to the bootstrap, Monographs and Statistics and Applied Probability, CRC press LLC, New-York, USA.

17. Fan R.E., Chen P.H., Lin C.J., 2005. Working set selection using second order information training SVM. J. Machine Learn. Res., 6, 889–1918.

18. Freund Y., Schapire R.E., 1996. Experiments with a new boosting algorithm. 13th International Conference on Machine Learning ICML'96, Bari, Italy, July 3–6.

19. Giacinto G., Roli F., 2001. Design of effective neural networks ensembles for image classification processes. Image Vision Computing J., 19, 699–707.

20. Grandvalet Y., Canu S., 2008. SVM and kernel methods matlab toolbox:ASI-INSA de Rouen. http://asi.insa-rouen.fr/enseignants/~arakoto/toolbox/, last accessed June 27 2014.

21. Guo L., Boukir S., 2013. Margin-based ordered aggregation for ensemble pruning. Pattern Recogn. Lett., 34, 603–609.

22. Hachichi M.S., Vahedian A., Yazdi H.S., 2011. Creating and measuring diversity in multiple classifier systems using support vector data description. Applied Soft Computing, 11, 4931–4942.

23. Han L., Han L., Liu C. 1999. Neural network applied to prediction of the failure stress for a pressurized cylinder containing defects. International Journal of Pressure Vessels and Piping, 76, 215-219

24. Han H.G., Qiao J.F., 2013. A structure optimisation algorithm for feedforward neural network construction. Neurocomputing, 99, 347–357

25. Hand D., Mannila H., Smyth P., 2001. Principles of data mining. The MIT press, Cambridge.

26. He X., Wang Z., Jin C., Zheng Y., Xue X., 2012. A simplified multi-class support vector machine with reduced dual optimization. Pattern Recogn. Lett., 33, 71–82.

27. Hernandez-Lobato D., Martinez-Munoz G., Suarez A., 2013. How large should ensembles of classifiers be? Pattern Recogn., 46, 1323–1336.

28. Ho T., 1998. The random subspace method for constructing decision forests. IEEE Transact. Pattern Anal. Machine Intell., 20, 8, 832–844.

29. Hsieh K.L., Tong L.I. 2001. Optimization of multiple quality responses involving qualitative and quantitative characteristics in IC manufacturing using neural networks. Computers in Industry, 46, 1-12.

30. Hsu S.C., Chien C.F., 2007. Hybrid data mining approach for pattern extraction from wafer bin map to improve yield in semiconductor manufacturing. International Journal of Production Economics, 107, 88-103

31. Hsu C.W., Chang C.C., Lin C.J., 2003. A practical guide to support vector classification. Technical Report, Department of Computer Science, National Taiwan University, http://www.csie.ntu.edu.tw/~cjlin, last accessed June 27 2014.

32. Hung Y.H. 2007. Optimal process parameters design for a wire bonding of ultra-thin CSP package based on hybrid methods of artificial intelligence. Microelectronics International, 24, 3-10

33. Ishikawa K., 1986. Guide to quality control. Asian Productivity Organization

34. Jain A.K, Murty M.N., Flynn P., 1999. Data clustering: a review. ACM Computing Surveys, 31, 264–323.

35. James G., Witten D., Hastie T., Tibshirani R., 2013. An Introduction to Statistical Learning. Springer, New York.

36. Kim Y.W., Oh I.S., 2008. Classifier ensemble selection using hybrid genetic algorithms. Pattern Recogn. Lett, 29, 796–802.

37. Kittler J., Hatef M., Duin R., Matas J., 1998. On combining classifiers. IEEE Transact. Pattern Anal. Machine Intell., 3, 226–239.

38. Ko A.H.R., Sabourin R., Britto A.S., 2008. From dynamic classifier selection to dynamic ensemble selection. Pattern Recogn., 41, 1718–1731.

39. Kohavi R., 1995. A study of cross-validation and bootstrap for accuracy estimation and model selection. Proc. of Int. Joint Conf. on Artificial Intelligence, San Mateo (USA), Morgan Kaufmann, 1137-1143.

40. Köksal G., Batmaz I., Testik M.C., 2011. A review of data mining applications for quality improvement in manufacturing industry. Expert Systems with Applications, 38, 13448-13467.

41. Kotsiantis, S.B., 2007. Supervised machine learning: a review of classification techniques. Informatica, 31, 249–268.

42. Kotsiantis S.B., 2011. Combining bagging, boosting, rotation forest and random subspace methods. Artif. Intell. Rev. 35, 225–240.

43. Kotsiantis S.B., 2013. Decision trees: a recent overview. Artif. Intell. Rev. 39, 261–283.

44. Krimpenis A., Bernardos P.G., Vosniakos G.C., Koukouvitaki A., 2006. Simulation-based selection of optimum pressure die-casting process parameters using neural nets and genetic algorithms. International Journal of Advances Manufacturing Technology, 27, 509-517

45. Kuncheva L.I., 2002. Switching between selection and fusion in combining classifiers: an experiment. IEEE Transact. Syst. Man Cybernetics, Part B: Cybernetics, 32, 2, 146–156.

46. Kuncheva L.I., 2004. Combining pattern classifiers: methods and algorithms. Wiley-Intersciences, Hoboken, USA, http://www.ccas.ru/voron/download/books/machlearn/kuncheva04combining.pdf, last accessed June 27 2014.

47. Kuncheva L.I., Whitaker C.J., 2003. Measures of diversity in classifier ensembles and their relationship with the ensemble accuracy. Machine Learning, 51, 181–207.

48. Kuncheva L.I., Whitaker C.J., Shipp C.A., 2003. Limits on the majority vote accuracy in classifier fusion. Pattern Anal. Appl., 6, 22–31.

49. Li A.D, He, Z., Zhang, Y., 2016. Bi-objective variable selection for key quality characteristics selection based on a modified NSGA-II and the ideal point method. Computers in Industry, 82, 95-103.

50. Li K., Liu Z., Han Y., 2012. Study of selective ensemble learning methods based on support vector machine. Physics Procedia, 33, 1518–1525.

51. Liang X., Ma Y., He Y., Yu L., Chen R.C., Liu T., Yang X., 2013. Fast pruning superfluous support vectors in SVMs. Pattern Recogn. Lett, 34, 1203–1209.

52. Mathworks, 2007. Statistics toolbox user's guide V 6.0. The MathWorks Inc., Natick

53. Mathworks, 2016. fileexchanges https://fr.mathworks.com/matlabcentral/fileexchange/58102-mlp-learning

54. Mehta M., Agrawal R., Rissanen J., 1996. SLIQ: A fast scalable classifier for data mining. Adv. Database Technol. EDBT '96, Lecture Notes in Computer Science, 1057, 18–32.

55. Meyer D., Leisch F., Hornik K., 2003. The support vector machine under test. Neurocomputing, 55, 169–186.

56. Nguyen D., Widrow B., 1990. Improving the learning speed of 2-layer neural networks by choosing initial values of the adaptive weights. Proc. of the Int. Joint Conference on Neural Networks IJCNN'90, 3, 21–26.

57. Noyel M., Thomas P., Charpentier P., Thomas A., Brault T. 2013. Implantation of an on-line quality process monitoring. 5th International Conference on Industrial Engineering and Systems Management IESM'13, Rabat, Maroc, 28 – 30 October.

58. Noyel M., Thomas P., Thomas A., Charpentier P.2016. Reconfiguration process for neuronal classification models: Application to a quality monitoring problem. Computers in Industry, 83, 78-91.

59. Oakland J.S., 2007. Statistical Process Control, Butterworth-Heinemann Ltd

60. Paliwal M., Kumar U.A., 2009. Neural networks and statistical techniques: a review of applications. Expert Systems with Applications, 36, 2–17.

61. Platt J.C., 1998. Sequential minimal optimization: a fast algorithm for training support vector machines. Technical Report MSR-TR-98-14, Microsoft Research.

62. Quinlan, J.R., 1996. Improved use of continuous attributes in C4.5. J. Artif. Intell. Res., 4, 77–90.

63. Rodriguez J.J., Kuncheva L.I., Alonso C.J., 2006. Rotation forest: a new classifier ensemble method. IEEE Trans. Pattern Anal. Mach. Intell., 28, 1619–1630.

64. Ruta D., Gabrys B., 2005. Classifier selection for majority voting. Information Fusion, 6, 63–81.

65. Santucci, E., Didaci, L., Fumera, G., Roli, F., 2017. A parameter randomization approach for constructing classifier ensembles. Pattern Recognition, 69, 1–13.

66. Shafer J., Agrawal R., Mehta M., 1996. SPRINT: a scalable parallel classifier for data mining. 22th International Conference on Very Large Data VLDB'96, Bombay, India September 3–6.

67. Soto V., Martinez–Munoz G., Hernadez–Lobato D., Suarez A., 2013. A double pruning algorithm for classification ensembles. 11[th] International Workshop on Multiple Classifier Systems MCS'13, Nanjing, China, May 15–17.

68. Standox, 2018. http://www.standox.com/content/dam/EMEA/Standox/HQ/Public/Documents/English/Standotheks/THK_Paint_Defects_GB.pdf

69. Taguchi G., 1989. Quality Engineering in Production Systems, NY: McGraw-Hill

70. Tang E.K, Suganthan P.N., Yao X., 2006. An analysis of diversity measures. Machine Learning, 65, 247–271.

71. Thomas P., Bloch G., Sirou F., Eustache V., 1999. Neural modeling of an induction furnace using robust learning criteria. J. Integrated Computer Aided Engineering, 6, 1, 5–23.

72. Thomas P., Noyel M., Suhner M.C., Charpentier P., Thomas A., 2013. Neural networks ensemble for quality monitoring. 5[th] International Joint Conference on Computational Intelligence IJCCI'13, Vilamoura, Portugal, September 20–22, 515–522.

73. Thomas P., Suhner M.C., 2015. A new multilayer perceptron pruning algorithm for classification and regression applications. Neural Processing Letters.42, 2, 437-458.

74. Thomas P., Thomas A., 2009. How deals with discrete data for the reduction of simulation models using neural network. 13[th] IFAC Symp. On Information Control Problems in Manufacturing INCOM'09, Moscow, Russia, June3–5, 1177–1182.

75. Tsoumakas G., Patalas I., Vlahavas I., 2009. An ensemble pruning primer, in:O. Okun, G. Valentini (Eds.), Applications of supervised and unsupervised ensemble methods. Studies in Computational Intelligence, Springer, London, UK.

76. Windeatt T., 2005. Diversity measures for multiple classifier system analysis and design. Information Fusion, 6, 21–36.

77. Wozniak M., 2007. Experiments with trained and untrained fusers, in E. Corchado, J. Corchado, A. Abraham (Eds.), Innovations in Hybrid Intelligent Systems, Advances in Soft Computing, 44, Springer, berlin 144–150.

78. Wozniak M., 2009. Evolutionary approach to produce classifier ensemble based on weighted voting,World Congress on Nature and Biologically Inspired Computing NaBIC'09, 648-653.

79. Wozniak M., Grana M., Corchado E., 2014. A survey of multiple classifier systems as hybrid systems. Information Fusion, 16, 3–17.

80. Xiaoqiao W, Mingzhou L., Maogen G., Lin L., Conghu L., (2015). Research on assembly quality adaptive control system for complex mechanical products assembly process under uncertainty. Computers in Industry, 74, 43-57.

81. Yang L., 2011. Classifiers selection for ensemble learning based on accuracy and diversity. Procedia Engineering, 15, 4266–4270.

82. Yang T., Tsai T., Yeh J., 2005. A neural network based prediction model for fine pitch stencil printing quality in surface mount assembly. Engineering Application of Artificial Intelligence, 18, 335-341

83. Yang X., Lu J., Zhang G., 2010. Adaptive pruning algorithm for least squares support vector machine classifier. Soft Computing, 14, 667–680.

84. Yu J., Xi L., Zhou X., (2008). Intelligent monitoring and diagnosis of manufacturing processes using an integrated approach of KBANN and GA. Computers in Industry, 59, 489-501

85. Zhou Z.H., Wu J., Tang W., 2002. Ensembling neural networks: many could be better than all. Artif. Intell., 137, 239–263.

86. Zhou X., Ma Y., Tu Y., Feng Y. 2013. Ensemble of surrogates for dual response surface modeling in robust parameter design. Quality and Reliability Engineering International, 29, 173-197.